\documentclass[journal]{IEEEtran}
\IEEEoverridecommandlockouts                              %
\usepackage{balance}
\usepackage{hyperref}
\usepackage{cite}
\usepackage{microtype}  %
\usepackage{graphicx}
\usepackage{amsmath}
\usepackage{afterpage}
\usepackage{amssymb}
\usepackage{booktabs}
\usepackage{textcomp}
\usepackage{multirow}
\usepackage[caption=false,font=footnotesize]{subfig}
\usepackage{makecell}
\usepackage{array}      

\begin{document}
\title{
Disturbance-Aware Motion Planning for Over-Actuated\\
Underwater Vehicles Exploiting Actuation Redundancy\\
for High-Fidelity 3D Reconstruction
}

\author{Yuer Gao, \IEEEmembership{Student Member, IEEE}, Tongqing Xu, Qingyang Liu, and Yi Cai, \IEEEmembership{Member, IEEE}%


\thanks{This work was supported by the Hong Kong University of Science and Technology (Guangzhou).}%

\thanks{$^{1}$All authors are with the Smart Manufacturing Thrust, Systems Hub, 
The Hong Kong University of Science and Technology (Guangzhou), Guangzhou, China.
{\tt\small yicai@hkust-gz.edu.cn}}%

\thanks{Digital Object Identifier (DOI): see top of this page.}
}


\maketitle
\thispagestyle{empty}
\pagestyle{empty}

\begin{abstract}

Underwater robots often operate near delicate targets where high-power thrusters resuspend sediments and induce turbulence, degrading image quality at the sensor input. Conventional controllers optimize vehicle-centric objectives (tracking and stability) without accounting for actuation’s impact on sensing. We address this actuation-to-perception coupling by exploiting redundancy in over-actuated platforms. For an eight-thruster ROV, multiple thrust allocations yield the same motion; we search this null space to minimize predicted disturbance in a task-relevant target region while enforcing motion constraints. Our method uses a control-oriented thruster-wake proxy derived from actuator-disk theory with directional attenuation and validated by PIV ($R^2 = 0.99$ near the wake axis; $R^2 > 0.82$ in the primary wake region), and a real-time redundancy-resolving allocator running at 10 Hz (45~ms/solve). Across 440 trials, the approach reduces target-region particle velocity by 67\% ($p < 0.001$), improves 3D reconstruction RMSE by 55\% versus a disturbance-unaware baseline ($1.9 \pm 0.4$~mm vs.\ $4.3 \pm 1.8$~mm), and achieves a 98.5\% reconstruction success rate. The framework supports autonomous scanning (quantitatively evaluated) and operator-assisted inspection (demonstrated in supplementary materials).

\end{abstract}

\begin{IEEEkeywords}
Over-actuated underwater vehicle, actuation redundancy, disturbance-aware control, thruster wake modeling, visual inspection.
\end{IEEEkeywords}

\section{Introduction}

\begin{figure}[ht]
  \centering
  \includegraphics[width=\linewidth]{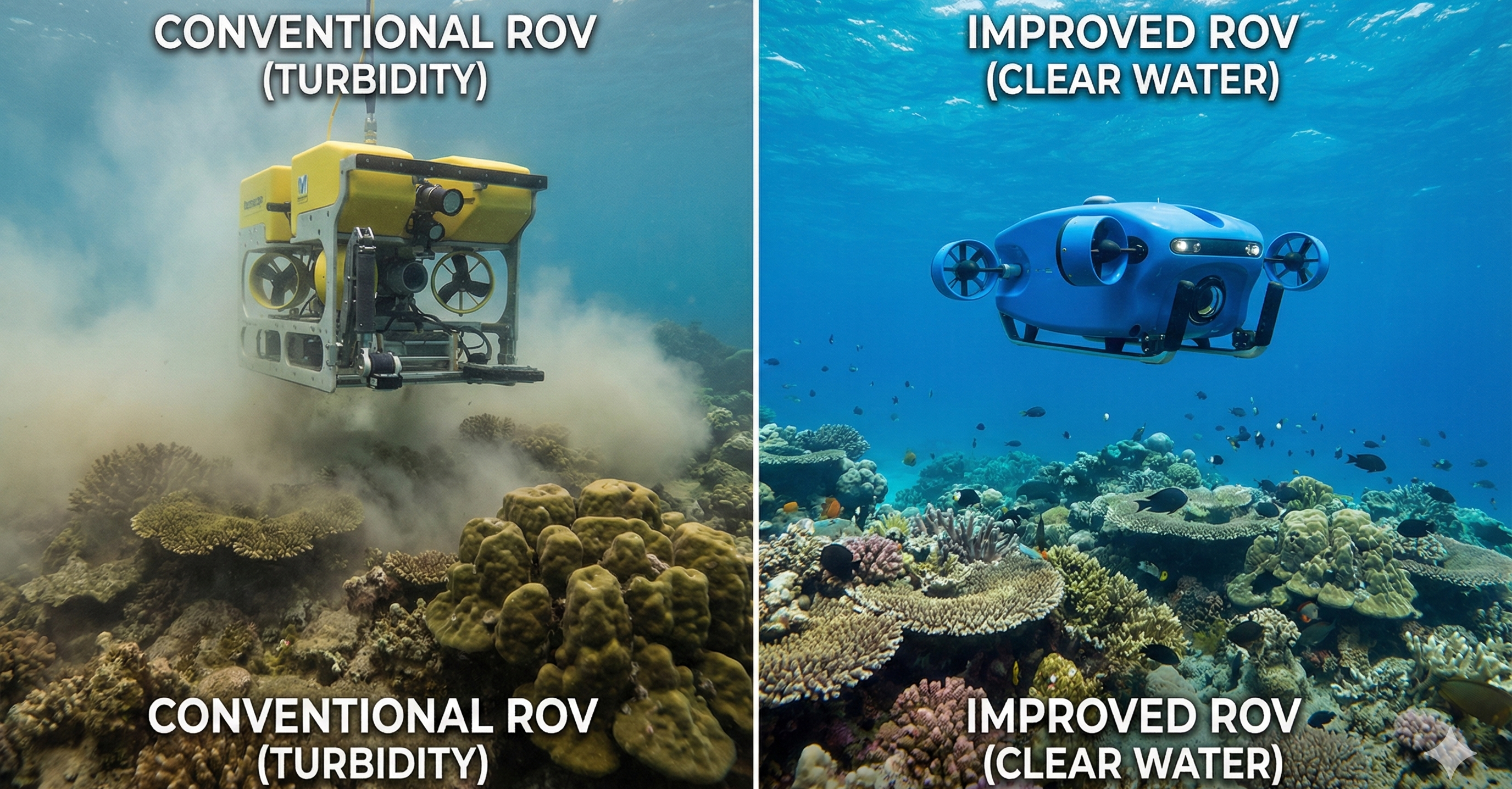}
  \caption{Thruster-induced disturbance comparison. (a) Conventional ROV 
  with sediment resuspension. (b) The ideal method maintains clear visibility.}
  \label{fig:disturbance_schematic}
\end{figure}

\IEEEPARstart{U}{nderwater} robots have become indispensable for infrastructure inspection, environmental monitoring, and scientific exploration. High-fidelity sensing of underwater structures is critical to these applications, from detecting hairline cracks in pipelines to monitoring coral bleaching and documenting archaeological sites. However, approaching a target creates a core problem: thruster actuation from the robots induces hydrodynamic disturbances that resuspend sediments~\cite{craig2023modeling, liao2015situ} and reduce water clarity (Fig.~\ref{fig:disturbance_schematic}), ultimately corrupting sensor data at the source\cite{li2018non}. This actuation-to-perception coupling—wherein a robot's thrusters degrade the data quality its sensors are trying to acquire—represents a mechatronic bottleneck that cannot be resolved through post-processing or algorithmic compensation alone.

Standard thruster configurations compound this problem. A conventional six-thruster underwater vehicle provides exactly six control inputs for six degrees of freedom—sufficient for motion control but with no redundancy for secondary objectives such as environmental disturbance minimization. Trajectory tracking and environmental disturbance thus become physically coupled: satisfying one constraint necessarily dictates the other. Reducing disturbance typically requires slower, gentler motions~\cite{chauhan2025investigation}, creating an apparent tradeoff between task efficiency and data quality. Without actuation redundancy, controllers cannot decouple these competing objectives.

Current motion controllers exemplify the first challenge: they optimize for vehicle-centric objectives while disregarding their environmental consequences. Trajectory tracking controllers minimize position error~\cite{fossen2011handbook}; dynamic positioning systems maximize station-keeping accuracy; robust controllers reject external disturbances acting on the robot~\cite{antonelli2014underwater}. None consider how the robot's actuation affects its surroundings. Although the sediment disturbance problem has been qualitatively documented—Liu et al.\cite{liu2024nature} observed thruster wakes impacting seabeds, and McLean et al.\cite{mclean2020enhancing} stressed the importance of avoiding downward-facing thrusters near sediments—no existing framework provides a quantitative model suitable for real-time control integration. This gap leaves controllers blind to the environmental impact of their commands.

Existing disturbance mitigation strategies operate at different system levels, each with inherent limitations. Hardware-level approaches—such as optimized propeller blade geometry \cite{carlton2018marine} or reconfigured thruster placement \cite{xu2024demand}—modify the disturbance characteristics of individual actuators but remain fixed once deployed and cannot adapt to varying inspection scenarios. Operational-level strategies reduce commanded velocity to limit disturbance indirectly 
\cite{chauhan2025investigation}, but at the cost of significantly prolonged task duration (often 3$\times$ or more). Algorithmic post-processing, including underwater image enhancement \cite{ancuti2012enhancing} and neural restoration \cite{levy2023seathru}, cannot recover information physically destroyed by motion blur or sediment occlusion.
In contrast, the proposed method operates at the control allocation level, where actuation redundancy enables real-time selection among infinitely many thrust distributions that achieve identical vehicle motion. This distinction is critical: redundancy resolution offers a compelling set of benefits—reducing disturbance while preserving motion authority, without relying on slower motions or task-specific hardware modifications.

Our approach differs fundamentally from prior methods: rather than compensating for degraded data after acquisition, we actively preserve scene integrity during the acquisition process itself. Over-actuation provides the redundancy needed to decouple motion tracking from disturbance minimization. An eight-thruster platform offers more control inputs than the six degrees of freedom required for motion, creating a null space of thrust configurations that achieve identical robot motion. Our framework exploits this null space to find solutions that minimize hydrodynamic disturbance on target regions while respecting operator intent. To this end, we develop a real-time thruster wake model that predicts flow velocity at arbitrary spatial locations, enabling disturbance to be formulated as an explicit optimization cost. Building on this model, we design a control allocation optimizer that solves the redundancy resolution problem at 10~Hz by selecting thrust allocations from the null space to minimize target disturbance. Finally, we integrate these components within a shared control architecture that interprets operator velocity commands and translates them into disturbance-optimized trajectories while preserving human authority over inspection strategy.

A key insight underlying our approach is the concept of \textit{gentle control}: achieving precise motion objectives through smooth, low-disturbance actuation rather than aggressive, high-bandwidth corrections. Traditional controllers optimize for \textit{tight stability}—minimizing position error through rapid thruster responses—but this creates local turbulence that undermines sensing objectives. Our framework instead pursues what we term \textit{gentle stability}—maintaining positioning accuracy while explicitly minimizing thruster-induced flow on target regions, in contrast to conventional tight stability that prioritizes tracking error regardless of environmental impact.

This paper makes three contributions. First, we derive a computationally efficient thruster wake model based on actuator disk theory with directional attenuation. Unlike prior qualitative descriptions~\cite{liu2024nature},~\cite{mclean2020enhancing}, our model is validated against Particle Image Velocimetry measurements ($R^2 = 0.84$) in the primary wake region and adds less than 1 ms computational overhead per control iteration, enabling explicit disturbance minimization as a real-time control objective. Building on this, we demonstrate that eight-thruster over-actuation provides a null space enabling simultaneous optimization of trajectory tracking and environmental preservation. To the best of our knowledge, this represents the first application of redundancy resolution principles to environmental disturbance minimization in underwater manipulation, achieving 67\% disturbance reduction without sacrificing positioning accuracy (2.1 ± 0.6~mm RMSE).

Finally, we develop a control allocation framework applicable to both autonomous and teleoperated modes. We validate the autonomous scanning mode quantitatively; teleoperation capability is demonstrated qualitatively in the supplementary video.
In autonomous mode, the system executes pre-planned viewpoint sequences with disturbance-aware redundancy-resolving allocation. In shared control mode, the architecture preserves operator decision-making authority for inspection strategy while automating low-level disturbance mitigation. Our experimental validation primarily demonstrates autonomous scanning performance, while Section~\ref{sec:control_framework} details the shared control mechanisms available for teleoperation scenarios.

\section{Related Work}

Our approach to the actuation-to-perception coupling problem draws on three intersecting research domains. We begin by reviewing underwater 3D reconstruction to establish the input quality requirements our control framework must satisfy, then examine underwater robot motion control to identify the gap between vehicle-centric objectives and acquisition-aware paradigms. A discussion of shared control approaches contextualizes our human-robot collaboration architecture. This intersection reveals that while individual components are well-studied, no existing framework explicitly optimizes actuation for downstream data quality.

\subsection{Underwater 3D Reconstruction}

Dedicated active optical scanners—including laser-based and structured-light systems~\cite{bruno2011experimentation, diamanti2021multi}—provide the highest accuracy for underwater 3D modeling, achieving millimeter-level precision. However, their cost and operational complexity make them impractical for routine inspection.

Photogrammetric reconstruction using multi-view imagery from low-cost ROVs offers a practical alternative. The reconstruction process follows a well-established pipeline: (1) image acquisition from multiple viewpoints, (2) feature extraction and matching across views, (3) camera pose estimation via bundle adjustment, and (4) dense geometry recovery. Structure-from-Motion (SfM) and Multi-View Stereo (MVS)~\cite{schonberger2016structure, furukawa2015multi} implement this pipeline using hand-crafted features and optimization-based depth estimation. Neural implicit representations—NeRF and 3D Gaussian splatting~\cite{kerbl20233d}—replace explicit geometry with learned volumetric or point-based representations, achieving superior fidelity for novel view synthesis.

Both classical and learning-based pipelines share strict input requirements. Feature matching demands sufficient texture and consistent appearance across views. Pose estimation requires overlap between adjacent frames and minimal motion blur. Dense reconstruction depends on accurate depth cues, which degrade under poor visibility. NeRF-based methods are particularly sensitive: training on blurred or occluded images produces artifacts that propagate throughout the reconstructed volume \cite{remondino2023critical}.

Underwater environments pose inherent challenges to each of these requirements. Color attenuation, backscatter, and nonuniform illumination degrade image clarity~\cite{li2018non}. Restoration algorithms such as Sea-thru~\cite{levy2023seathru} partially compensate for optical effects but cannot recover information destroyed by motion blur or sediment occlusion. When a thruster disturbs the water column, suspended particles scatter light and obscure the target—damage that no post-processing can reverse. This identifies a practical constraint: reconstruction quality is bounded at the source by the image acquisition process, not by downstream algorithmic capability.

\subsection{Underwater Robot Motion Control}

Underwater vehicle control research addresses vehicle stability and trajectory adherence through several approaches. Systematic multi-view coverage requires accurate path following to ensure sufficient overlap between captured frames. Classical approaches use PID controllers ~\cite{fossen2011handbook}, while advanced methods employ Model Predictive Control (MPC)~\cite{shen2016integrated} or Sliding Mode Control (SMC)~\cite{joe2014second} to handle nonlinear hydrodynamics. These controllers optimize tracking error but do not consider how actuation affects image quality during the trajectory.

Station-keeping is essential for capturing sharp images at fixed viewpoints. Dynamic Positioning (DP) systems maintain pose against environmental forces~\cite{veksler2016dynamic} and achieve centimeter-level accuracy. However, tight station-keeping requires high-bandwidth thruster corrections that generate local turbulence—potentially degrading the imagery that stable positioning was intended to enable.

Underwater robots operate amid currents, waves, and tether drag. Robust control strategies—$H_\infty$control~\cite{katebi1997h}, adaptive control ~\cite{guerrero2020adaptive}, and Disturbance Observer-Based Control (DOBC) ~\cite{chen2015disturbance}—maintain stability under these conditions. These methods reject disturbances \textit{acting on} the robot but do not address disturbances \textit{caused by} the robot's own thrusters.

Real-time planning on embedded systems uses sampling-based planners (RRT variants) or optimization-based methods ~\cite{shen2016integrated}. These planners optimize time, energy, or smoothness. None incorporates the effect of planned motions on surrounding water clarity or target visibility.

Beyond control approaches, disturbance could theoretically be addressed through hardware modifications (specialized low-wash thrusters, relocated thruster placement) or operational constraints (reduced speed, wait-and-settle protocols). However, hardware solutions increase cost and complexity, while operational constraints extend mission duration beyond practical limits. Over-actuation offers a unique advantage: it exploits the mathematical redundancy inherent in the thruster configuration to minimize disturbance without sacrificing motion capability or requiring specialized hardware, making it particularly suitable for low-cost inspection platforms.

\subsection{Shared Control for Teleoperation}

Low-cost multi-thruster ROVs have expanded access to underwater environments for inspection and data collection. Direct teleoperation remains difficult: operators face high cognitive workloads and struggle with precise 6-DOF maneuvering during close-range inspection~\cite{nauert2023inspection}.

Shared control combines human judgment with autonomous assistance. Existing methods use visual servoing for positioning~\cite{chaumette2006visual}, potential fields for obstacle avoidance~\cite{rosenberg1993virtual}, or predefined trajectories for navigation~\cite{shek2022learning}. These approaches refine human input for feasibility or safety but do not account for the data quality implications of the resulting motion.

For multi-view reconstruction, operators must guide the robot through a sequence of viewpoints while maintaining image quality. Self-induced hydrodynamic disturbance directly undermines this goal, as abrupt thruster inputs create turbulence that stirs sediment and corrupts the images being collected. Aggressive counter-actuation for disturbance rejection can worsen local turbulence. No existing shared control framework treats minimization of thruster-induced disturbance as a control objective tied to downstream reconstruction quality.


Table~\ref{tab:comprehensive_comparison} compares control objectives across methods. Existing controllers optimize vehicle-centric metrics: tracking accuracy, stability, and disturbance rejection. Reconstruction methods assume clean input images. The connection between robot motion and acquired data quality remains unaddressed.
Table~\ref{tab:related_methods} provides a systematic comparison of our approach with related methods. Unlike prior work that either ignores self-induced disturbance or addresses it only qualitatively, our framework explicitly models thruster-induced hydrodynamic effects and integrates real-time disturbance minimization with shared control.

This work closes the loop by introducing acquisition-aware control. We formulate thruster-induced disturbance as an explicit cost and leverage platform over-actuation to decouple motion objectives from environmental impact. This enables low-cost ROVs to execute the gentle, precise maneuvers required to capture high-quality multi-view imagery for 3D reconstruction.

\begin{table*}[t] 
\centering
\caption{Comparison with Major Methods: Methodological Capabilities and Control Objectives}
\label{tab:comprehensive_comparison}
\renewcommand{\arraystretch}{1.2} 
\setlength{\tabcolsep}{5pt} 
\begin{tabular}{lcccccc}
\toprule
\multirow{2}{*}{\textbf{Method Category}} & \textbf{Disturbance} & \textbf{Real-Time} & \textbf{Primary} & \textbf{Tracking} & \textbf{Env.} & \textbf{Recon.} \\
 & \textbf{Strategy} & \textbf{Optimization} & \textbf{Objective} & \textbf{Accuracy} & \textbf{Prot.} & \textbf{Ready} \\
\midrule
PID/DP Control \cite{fossen2011handbook} & Ignored & \textbf{Yes} & Stability & High & Low & No \\
Robust Control \cite{fossen2011handbook} & \textbf{Rejection} & \textbf{Yes} & Stability & High & Low & No \\
MPC Planning \cite{shen2016integrated} & Rejection & No/Limited$^{\dag}$ & Optimality & \textbf{High} & Low & Low \\
Shared Control \cite{chaumette2006visual} & Rejection & \textbf{Yes} & Usability & Med & Med & Med \\
\midrule
\textbf{Ours} & \textbf{Minimization} & \textbf{Yes} & \textbf{Data Quality} & \textbf{High} & \textbf{High} & \textbf{Yes} \\
\bottomrule
\end{tabular}
\\
\raggedright 
\footnotesize{$^{\dag}$ Standard MPC often struggles with high-frequency updates on embedded hardware. \textit{Recon. Ready}: Suitability for high-fidelity 3D reconstruction tasks. \textit{Env. Prot.}: Environmental Protection.}
\end{table*}

\begin{table*}[t]
\centering
\caption{Comparison of Related Methods}
\label{tab:related_methods}
\begin{tabular}{lcccccc}
\toprule
\textbf{Method} & \textbf{Year} & \textbf{Disturbance} & \textbf{Real-Time} & \textbf{Reconstruction} & \textbf{Shared} & \textbf{Platform} \\
 & & \textbf{Modeling} & \textbf{Optimization} & \textbf{Integration} & \textbf{Control} & \\
\midrule
Casalino et al. \cite{rist2016autonomous} & 2016 & None & Yes & No & Yes & 6-DOF AUV \\
Ridao et al. \cite{ridao2015intervention} & 2015 & None & Yes & No & Yes & I-AUV \\
Kapoutsis et al. \cite{kapoutsis2016real} & 2022 & External only & Yes & No & No & 6-thruster \\
Schmid et al. \cite{schmid2020efficient} & 2020 & None & Yes & Yes & No & UAV \\
Levy et al. \cite{levy2023seathru} & 2023 & Implicit & No & Yes & No & Diver-held \\
Shukla \& Karki \cite{shukla2016application} & 2016 & Qualitative & No & No & No & ROV \\
\midrule
\textbf{Ours} & \textbf{2026} & \textbf{Explicit} & \textbf{Yes} & \textbf{Yes} & \textbf{Yes} & \textbf{8-thruster} \\
\bottomrule
\end{tabular}
\end{table*}

\begin{figure*}[t]
    \centering
    \includegraphics[width=0.85\textwidth]{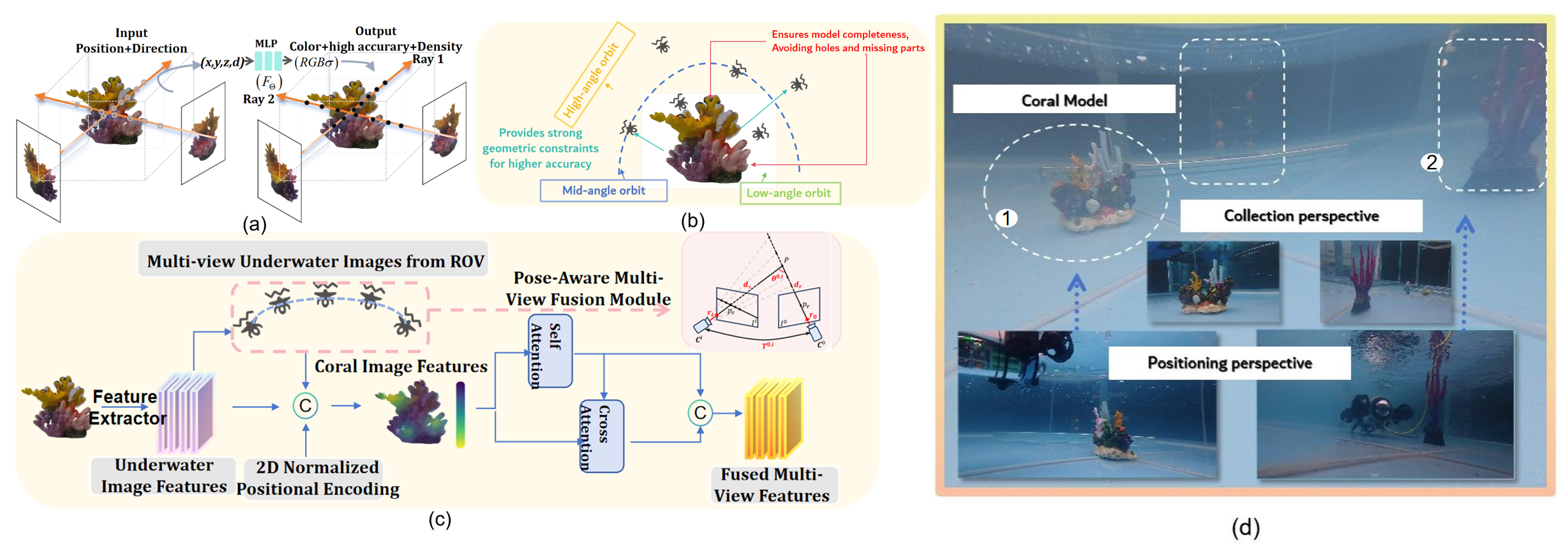}
    \caption{Overview of the proposed framework and evaluation pipeline.
(a) Reconstruction-based evaluation pipeline (used for evaluation only).
(b) Hemispherical layered-orbit scanning strategy.
(c) Standard multi-view reconstruction pipeline on acquired images.
(d) Tank experimental setup with target regions.}
    \label{fig:methodology_overview}
\end{figure*}

\section{METHODOLOGY}

Central to our approach is a disturbance-aware motion commander that orchestrates high-fidelity underwater 3D reconstruction. This planner uses a custom-designed, over-actuated robotic platform to generate trajectories that actively minimize hydrodynamic disturbance on targets. We first introduce the thruster disturbance model (Section~\ref{sec:system_overview}), then describe the viewpoint sampling strategy (Section~\ref{sec:disturbance_model}), followed by the redundancy-resolving allocation formulation (Section~\ref{sec:control_framework}) and hardware implementation (Section~\ref{sec:hardware_platform}). The 3D reconstruction validation pipeline as shown in Fig.~\ref{fig:methodology_overview}---comprising multi-view image acquisition, feature extraction, and surface reconstruction---is detailed in Section~\ref{sec:experimental_validation} alongside experimental results.

\subsection{System Overview}
\label{sec:system_overview}
\begin{figure*}[t]
\centering
\includegraphics[width=0.75\textwidth]{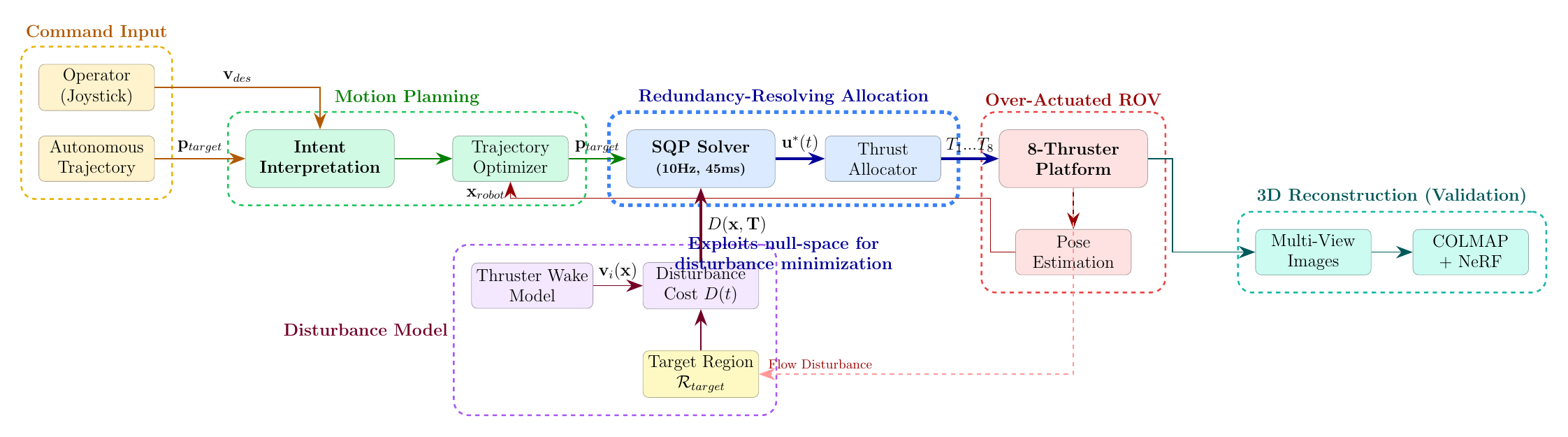}
\caption{System architecture of the disturbance-aware shared control 
framework. The human operator provides velocity commands $\mathbf{v}_{des}$ 
via a joystick, which is interpreted and optimized by the real-time motion 
planner. The SQP solver (10 Hz, 45 ms average) computes optimal thrust 
allocation $\mathbf{u}^*(t)$ by minimizing the disturbance cost $D(\mathbf{x},T)$ 
evaluated over the target region $R_{target}$. The over-actuated 8-thruster 
platform provides the redundancy necessary to decouple motion tracking 
from disturbance minimization.}
\label{fig:system_arch}
\end{figure*}

\begin{figure}[ht]
    \centering
    \includegraphics[width=\columnwidth]{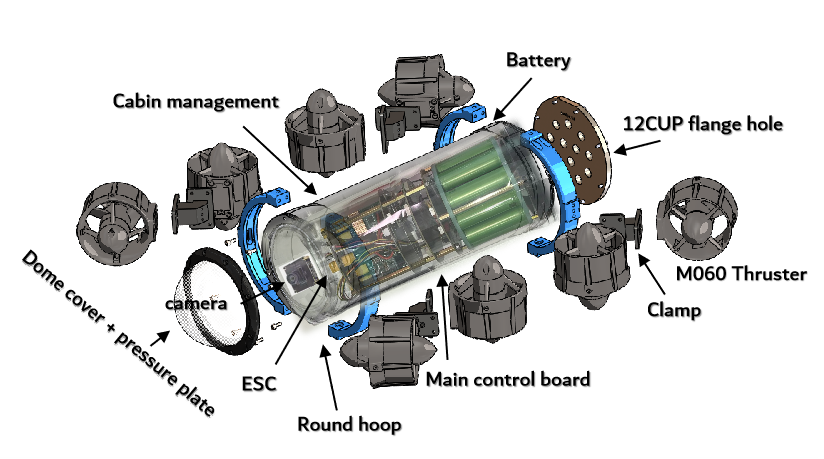} 
    \caption{The mechanical design and exploded view of our custom-built eight-thruster underwater robot, detailing key components.}
    \label{fig:hardware_overview}
    \vspace{-3mm} 
\end{figure}

\begin{figure}[ht]
    \centering
    \includegraphics[width=0.85\columnwidth]{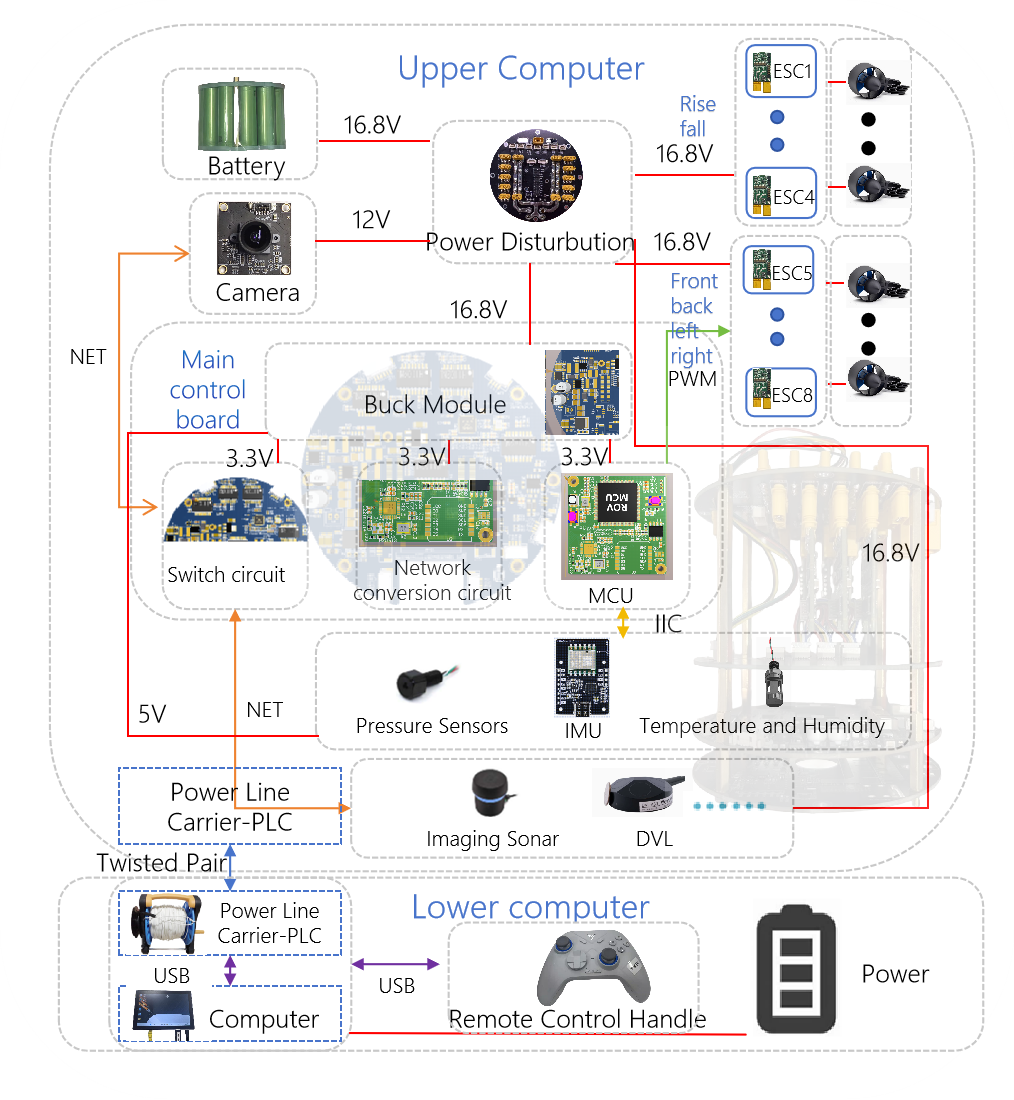}
    \caption{The electrical and communication architecture of the robotic system, illustrating the power distribution and data flow.}
    \label{fig:electrical_system}
    \vspace{-3mm} 
\end{figure}

Our framework (Fig.~\ref{fig:system_arch}) consists of: (1) an over-actuated eight-thruster underwater robot providing full 6-DOF control with redundancy, as shown in the mechanical design in Fig.~\ref{fig:hardware_overview}, and the electrical and communication architecture of the robotic system(Fig.~\ref{fig:electrical_system}); (2) a disturbance model predicting hydrodynamic forces on targets; (3) a real-time control allocation optimizer minimizing disturbance while respecting operator intent; (4) a low-level controller executing optimized commands.

\begin{table}[t]
\centering
\caption{Hardware Platform Specifications}
\label{tab:hardware_ultracompact}
\setlength{\tabcolsep}{3pt}
\renewcommand{\arraystretch}{0.88}
\begin{tabular}{@{}p{0.28\linewidth}p{0.67\linewidth}@{}}
\toprule
\textbf{Item} & \textbf{Specification} \\
\midrule
Vehicle & 450$\times$350$\times$280\,mm; 4.5\,kg; 30\,m depth rating \\
Propulsion & 8$\times$M060; vectored (4H/4V); 51\,N fwd / 40\,N rev (each) \\
Sensing & 4K@30\,fps (90$^\circ$ FOV); IMU 50\,Hz (JY901); MS5837  \\
\bottomrule
\end{tabular}
\end{table}

\begin{figure}[t]
  \centering
  \includegraphics[width=\linewidth]{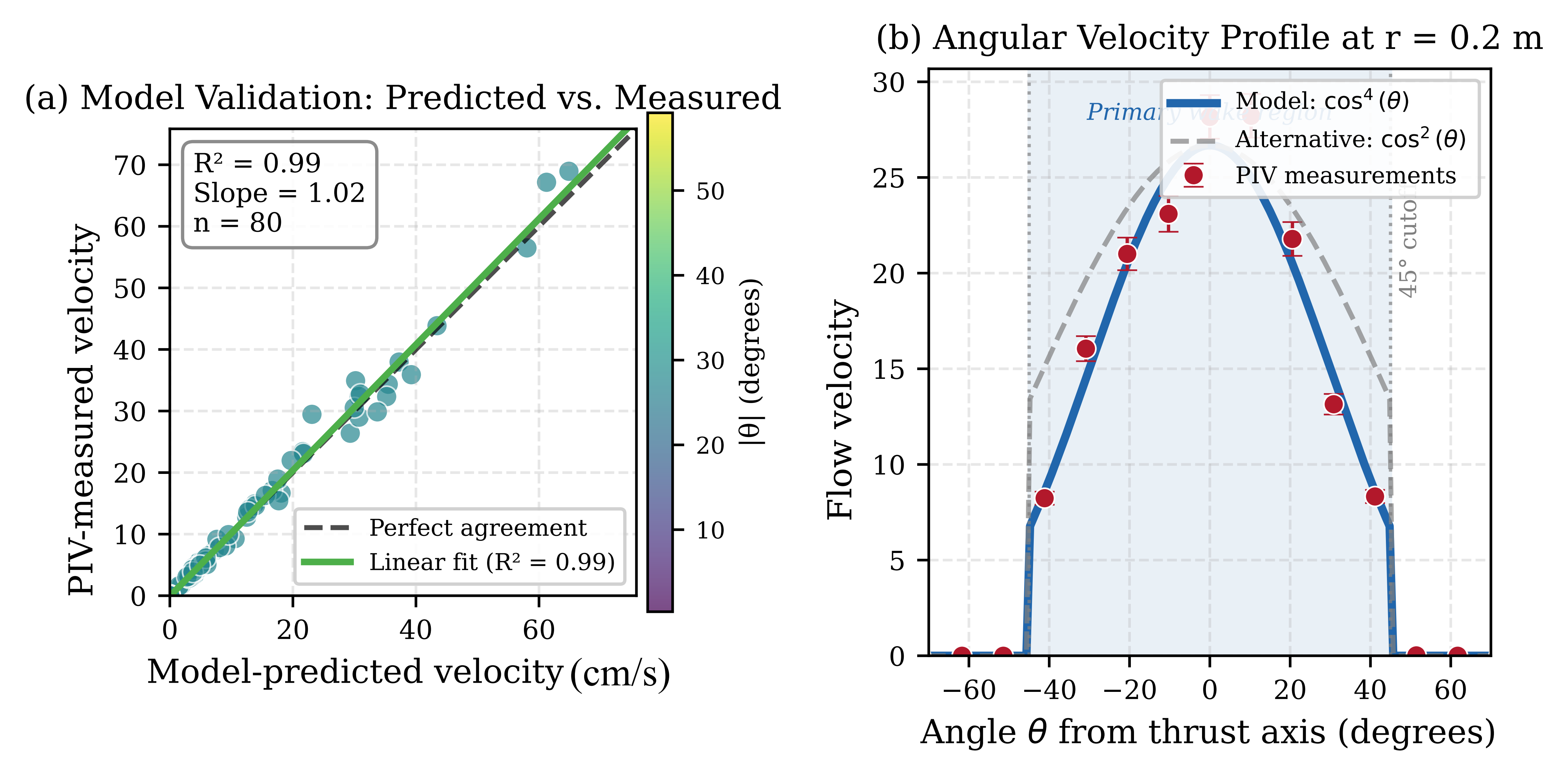}
  \caption{Validation of disturbance model. (a) Predicted vs. measured 
  velocities ($R^2 = 0.99$). (b) Angular profile showing $\cos^4(\theta)$ 
  fit vs. PIV data at $r = 0.2$~m.Measurements were conducted at distances of 0.1–0.5~m from the thruster 
  center, where axial velocities ranged from approximately 5–60 cm/s depending on thrust level and angular position.}
  \label{fig:piv_validation}
\end{figure}

\begin{figure*}[t]
    \centering
    \includegraphics[width=0.55\textwidth]{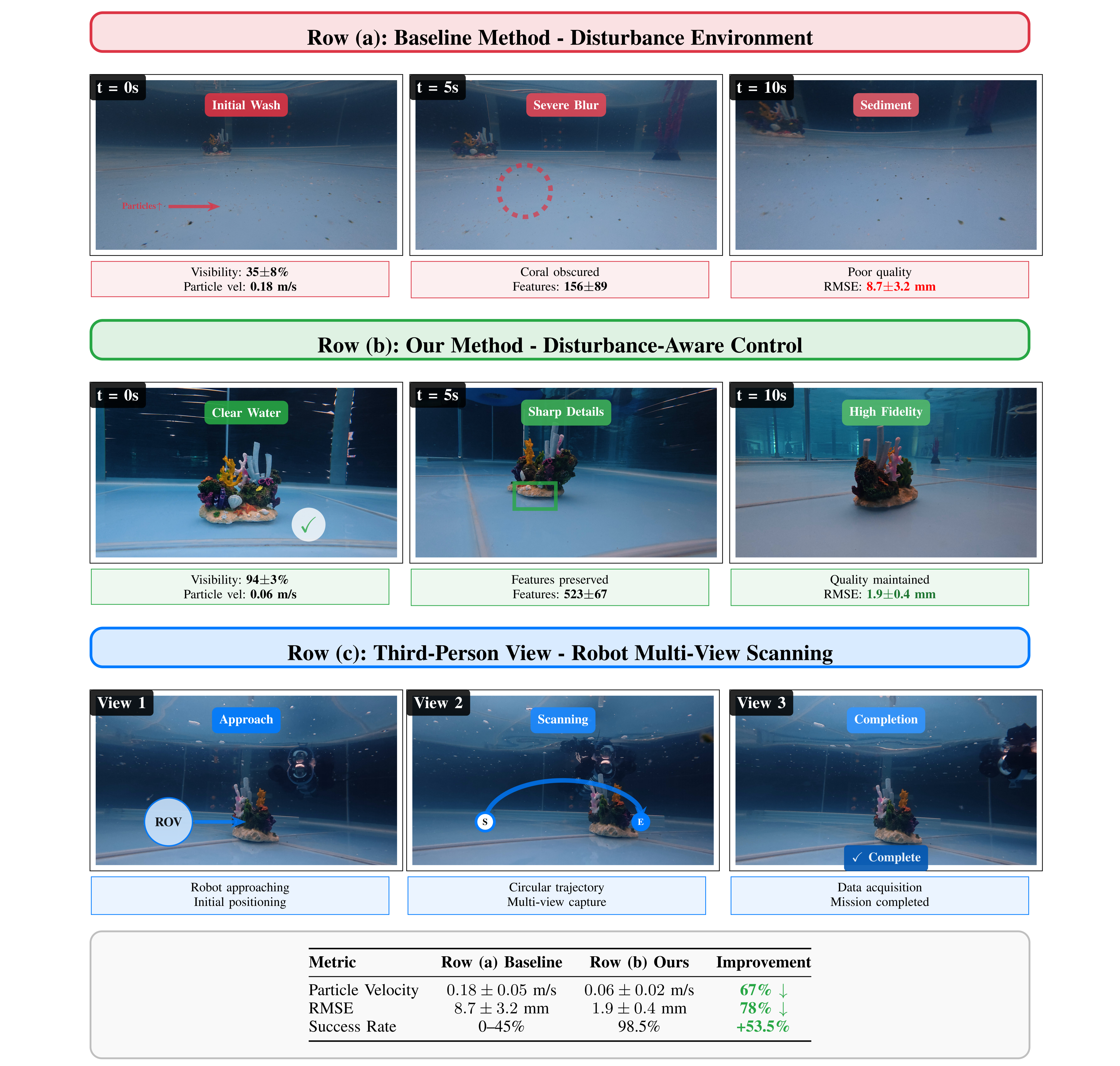}
    \caption{%
\textbf{Temporal comparison of water quality and reconstruction performance under different control strategies.} 
\textbf{Row (a):} Baseline method (manual/disturbance-Unaware) showing progressive water disturbance at three time points ($t=0$s, 5s, 10s), resulting in suspended particles, motion blur, and poor reconstruction quality (RMSE: $8.7 \pm 3.2$ mm). 
\textbf{Row (b):} Our disturbance-aware method maintaining clear water and sharp image quality throughout the scanning process, enabling high-fidelity reconstruction (RMSE: $1.9 \pm 0.4$ mm, 78\% improvement). 
\textbf{Row (c):} Third-person view showing the robot's multi-view scanning process from approach to completion, demonstrating the planned trajectory execution. 
Scale bars: 5 cm in all images. 
}
    \label{fig:disturbance_comparison}
    \vspace{-3mm} 
\end{figure*}

\begin{figure*}[t]
    \centering
    \includegraphics[width=0.45\textwidth]{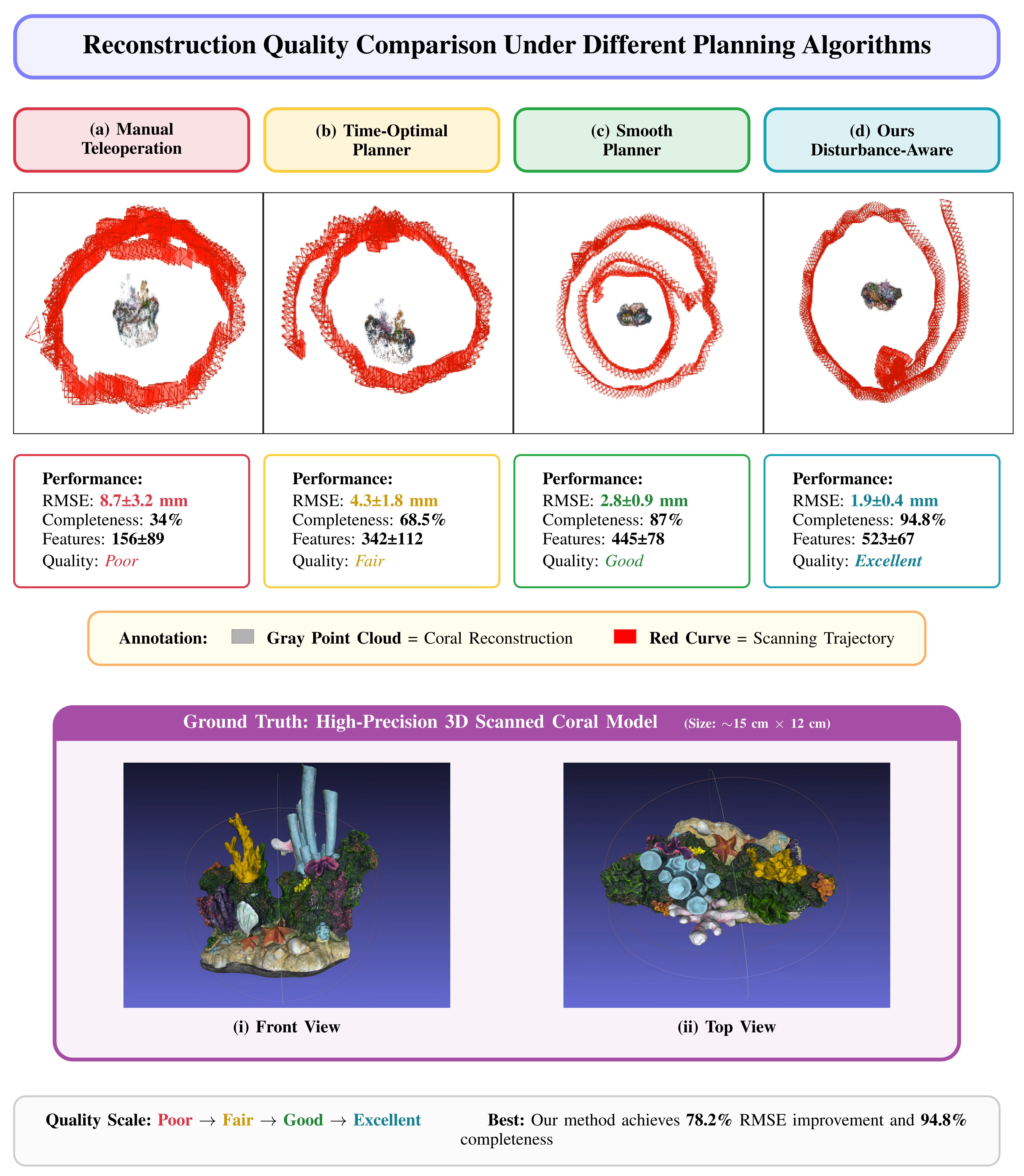}
    \caption{%
Comprehensive comparison of reconstruction quality under different motion planning algorithms. 
(a) Manual teleoperation results in chaotic trajectories with poor coverage. 
(b) Disturbance-Unaware planner improves efficiency but sacrifices quality (RMSE: 4.3$\pm$1.8 mm, 68.5\% completeness). 
(c) Smooth planner achieves good quality with refined trajectories (RMSE: 2.8$\pm$0.9 mm, 87\% completeness). 
(d) Our disturbance-aware method achieves the best reconstruction quality with systematic coverage (RMSE: 1.9$\pm$0.4 mm, 94.8\% completeness, 78.2\% improvement vs. baseline). 
All experiments were conducted on the same coral model ($\sim$15 cm $\times$12 cm).
}
\label{fig:reconstruction_comparison}
    \vspace{-3mm} 
\end{figure*}

\subsection{Disturbance Modeling}
\label{sec:disturbance_model}
\subsubsection{Physical Foundation}

We model each thruster as an idealized actuator disk~\cite{hou2019hydrodynamic} generating flow velocity at a spatial location $\mathbf{x}$, which approximates each thruster as a momentum source inducing velocity perturbations in the surrounding fluid. Unlike traditional applications of actuator disk theory focused on thrust generation, our model emphasizes the \textit{downstream disturbance field} affecting nearby targets:
\begin{equation}
\mathbf{v}_i(\mathbf{x}) = \underbrace{\sqrt{\frac{T_i}{2\pi\rho A_{prop}}}}_{\text{axial velocity}} 
                  \cdot \underbrace{\frac{r_{ref}^2}{r_i^2}}_{\text{distance decay}} 
                  \cdot \mathbf{n}_i \cdot g(\theta_i),
\label{eq:single_thruster}
\end{equation}
where $T_i$ is the thrust magnitude of the $i$-th thruster (N), $\rho=998$~kg/m$^3$ (water density), $A_{\text{prop}}$ is propeller disk area. the term $r_{ref}^2/r_i^2$ represents the inverse-square decay of velocity with distance from the thruster, $r_i = \|\mathbf{x} - \mathbf{x}_i\|$ is distance from the thruster center, normalized by the reference distance $r_{ref} = 1$~m for dimensional consistency. Note that $\mathbf{n}_i \in \mathbb{R}^3$ is the unit vector along the thrust direction, making $\mathbf{v}_i$ a vector quantity representing both magnitude and direction of induced flow.

The directional attenuation function captures jet-like flow patterns:
\begin{equation}
g(\theta_i) = \begin{cases}
\cos^4(\theta_i) & \text{if } |\theta_i| \leq 45^\circ \\
0 & \text{otherwise}
\end{cases},
\label{eq:directional_attenuation}
\end{equation}
where $\theta_i$ is the angle between $(\mathbf{x} - \mathbf{x}_i)$ and $\mathbf{n}_i$.

The directional attenuation function $g(\theta_i) = \cos^4(\theta_i)$ derives from established propeller wake theory. According to momentum theory for ducted propellers~\cite{carlton2018marine}, the axial velocity in a propeller jet decays approximately as $\cos^2(\theta)$ due to radial spreading of the momentum flux. Since the hydrodynamic force on a target scales with dynamic pressure (proportional to velocity squared), the disturbance effect follows a $\cos^4(\theta)$ distribution. This formulation is consistent with experimental observations by Lam et al.~\cite{sun2020numerical}, 
who measured thruster wake profiles showing similar angular decay patterns in underwater vehicle applications.

The 45° cutoff angle corresponds to the half-angle of the primary momentum flux cone observed in submerged jet studies~\cite{albertson1950diffusion}. Beyond this angle, the induced velocity drops below 10\% of the axial peak 
value, contributing negligibly to target disturbance. This simplification enables real-time computation while capturing the dominant disturbance physics.

\textbf{Model Validation:} To verify our simplified model, we compared predicted flow velocities against Particle Image Velocimetry (PIV) measurements at distances of 0.1--0.5~m from the thruster (Fig.~\ref{fig:piv_validation}). 

The model achieves strong agreement with PIV measurements: $R^2 = 0.99$ in the axial core region ($|\theta| < 15^\circ$,$r < 0.3$~m), where the momentum jet assumption holds well, decreasing to $R^2 = 0.82$ when including the full primary wake region ($|\theta| < 30^\circ$), where turbulent mixing introduces additional variance. The model tends to slightly overestimate disturbance at far distances ($r > 0.4$~m), providing a conservative bound suitable for real-time optimization where the objective is relative disturbance minimization rather than absolute flow prediction.

\subsubsection{Aggregate Disturbance Metric}

For an eight-thruster vehicle, total flow disturbance at target region $\mathcal{R}_{\text{target}}$ is:
\begin{equation}
D(\mathbf{x}_{\text{robot}}, \mathbf{T}) = \int_{\mathcal{R}_{\text{target}}} \left\| \sum_{i=1}^{8} \mathbf{v}_i(\mathbf{x}) \right\|^2 d\mathbf{x},
\label{eq:disturbance_integral}
\end{equation}

For real-time computation, we discretize using $N$ sample points $\{\mathbf{p}_j\}_{j=1}^N$ weighted by proximity to delicate features:
\begin{equation}
D(\mathbf{x}_{\text{robot}}, \mathbf{T}) \approx \sum_{j=1}^{N} w_j \left\| \sum_{i=1}^{8} \mathbf{v}_i(\mathbf{p}_j) \right\|^2,
\label{eq:aggregate_discrete}
\end{equation}
where $w_j$ are spatial weights (higher near intricate structures). We use $N=150-300$ points via stratified sampling (surface + volume), adding $<$1~ms computation per iteration.Minimizing $D(t)$ operationalizes our gentle control philosophy: among the infinitely many thrust allocations achieving identical motion, we select those inducing minimal hydrodynamic disturbance on the target region. This enables what we term gentle stability—precise positioning without the turbulent wake characteristic of high-gain conventional controllers.

Modeling assumptions: in~(\ref{eq:aggregate_discrete}), linear superposition of induced velocities is assumed for computational efficiency. While multi-jet interactions can be nonlinear, our objective requires a reliable relative ordering of disturbance across candidate allocations rather than exact flow prediction. Linear superposition tends to overestimate disturbance when wakes partially cancel, yielding a conservative bias appropriate for protection-oriented objectives. The 45$^\circ$ cutoff is chosen to exclude directions where measured induced velocities fall below 10\% of the axial peak in our PIV data (Fig.~\ref{fig:piv_validation}).

The number of sample points $N$ is determined by target complexity: $N = 150$ for simple geometries (Model C), $N = 300$ for complex branching structures (Model A). Weights $w_j$ are computed as:
\begin{equation}
w_j = \exp(-d_j/\sigma),
\label{eq:weights}
\end{equation}
where $d_j$ is the distance to the nearest delicate feature and $\sigma=0.05$ m is a spatial decay constant corresponding to approximately one-third of the typical coral branch spacing in our target models.

\subsection{Shared Control Architecture}
\label{sec:control_framework}
Manual control offers limited practical relief. Teleoperation of six-degree-of-freedom underwater vehicles is inherently demanding~\cite{wu20186}. Operators must simultaneously manage vehicle positioning, camera orientation, and task execution while avoiding obstacles. Expecting them to additionally modulate approach velocity and thruster commands to minimize sediment disturbance exceeds practical cognitive limits. Even experienced operators struggle to achieve the gentle, precise trajectories required for high-quality data acquisition near delicate targets.

Our framework supports two operational modes, though experimental validation in Section~\ref{sec:experimental_validation} focuses on autonomous scanning to enable controlled comparison of trajectory planning algorithms:

\textbf{Autonomous Scanning Mode (Primary):} The system executes pre-planned hemispherical trajectories with disturbance-aware optimization. This mode is used for systematic multi-view data acquisition and constitutes the main experimental evaluation.

\textbf{Assisted Teleoperation Mode (Framework Capability):} The operator provides velocity commands via joystick; the system optimizes thrust allocation to achieve commanded motion while minimizing disturbance. This mode demonstrates the framework's flexibility for exploratory inspection scenarios and is illustrated qualitatively in the supplementary video.

Our shared control framework implements a \emph{mediated teleoperation} paradigm~\cite{dragan2013policy}, where the operator provides high-level motion intent through velocity commands $\mathbf{v}_{des}$, and the system 
autonomously optimizes the low-level thrust allocation to achieve this motion while minimizing environmental disturbance.

\subsubsection{Intent-Disturbance Tradeoff}
The core insight is that over-actuation provides a null-space of thrust configurations achieving identical robot motion. Our optimizer searches this null-space to find solutions minimizing disturbance cost $D(t)$ while respecting the operator's intent.

The weight $w_{track}$ in~\eqref{eq: optimization} explicitly mediates the tradeoff between operator intent fidelity and disturbance minimization:
\begin{itemize}
    \item \textbf{High $w_{track}$} ($>10$): Prioritizes strict following of operator velocity commands, suitable for navigation phases where speed matters more than environmental preservation
    \item \textbf{Low $w_{track}$} ($<2$): Prioritizes gentle stability—allowing larger trajectory deviations to minimize target disturbance, suitable for delicate scanning phases where data quality is paramount
\end{itemize}
In our implementation, $w_{track} = 5.0$ was empirically tuned through grid search ($w_{track} \in \{1, 2, 5, 10, 20\}$) to achieve $<15\%$ deviation from operator-commanded velocity while reducing target disturbance by $>60\%$ compared to direct thrust allocation without optimization.

\subsubsection{Optimization Formulation}

Given operator command $\mathbf{v}_{des}$ (desired velocity in body frame), we solve the following redundancy-resolving allocation problem:

\begin{align}
\min_{\mathbf{p}(t),\mathbf{u}(t)} \quad & \int_0^T \Big( w_D \cdot D(t) + w_u \|\mathbf{u}(t)\|^2 \Big) dt + w_T \cdot T \nonumber \\
& + w_{track} \|\mathbf{p}(T) - \mathbf{p}_{target}\|^2 \label{eq:optimization} \\
\text{s.t.} \quad & \dot{\mathbf{x}}_{robot} = f(\mathbf{x}_{robot}, \mathbf{u}) \nonumber \\
& \mathbf{u}_{min} \leq \mathbf{u}(t) \leq \mathbf{u}_{max} \nonumber \\
& \|\dot{\mathbf{u}}(t)\| \leq \dot{u}_{max} \quad \text{(rate limits)} \nonumber \\
& \|\mathbf{x}(t) - \mathbf{x}_{obs}\| \geq d_{\text{safe}} \quad \text{(collision)},
\label{eq: optimization}
\end{align}
where $\mathbf{p}(t) \in \mathbb{R}^3$ denotes the robot position (translational component of the full state $\mathbf{x}_{robot} = [\mathbf{p}^\top, \boldsymbol{\Theta}^\top]^\top$ with orientation $\boldsymbol{\Theta} \in SO(3)$), and $\mathbf{p}_{target} = \mathbf{p}_{current} + \mathbf{v}_{des} \cdot T$ converts the 
operator's instantaneous velocity command to a soft position target over horizon $T$. The disturbance cost  $D(t)$ is disturbance cost, $T$ is the horizon duration, and $\mathbf{u}$ is the thrust vector.

The central observation is that overactuation provides multiple solutions to achieve $\dot{\mathbf{x}}_{\text{robot}}$. By exploiting this redundancy, our optimizer selects controls that minimize D(t) while respecting motion goals
\subsubsection{Real-Time Implementation}

\textbf{Handling Non-Convexity:} The disturbance cost $D(t)$ is 
non-convex due to the quadratic velocity superposition in~(\ref{eq:aggregate_discrete}) and the complex dependence on 
robot pose through thruster positions $\mathbf{x}_i$. We employ 
two strategies to obtain good solutions efficiently:
\begin{enumerate}
    \item \emph{Local Linearization}: At each SQP iteration, 
          $D(t)$ is approximated by its second-order Taylor expansion 
          around the current solution, yielding a convex QP subproblem.
    \item \emph{Warm-Starting}: Solutions from the previous control 
          cycle (at 10~Hz) initialize the current optimization. Given 
          the temporal continuity of robot motion and operator commands, 
          consecutive problems share similar optimal solutions, allowing 
          the solver to converge within the local basin of the previous 
          optimum.
\end{enumerate}

\textbf{Constraint Handling:}
Actuator limits ($\mathbf{u}_{min} \le \mathbf{u} \le \mathbf{u}_{max}$) are enforced as hard inequality constraints using an active-set strategy. Rate limits ($\|\dot{\mathbf{u}}\| \le \dot{u}_{max}$) are implemented as soft constraints with a quadratic penalty ($\rho=100$) to allow brief violations during aggressive maneuvers. Collision avoidance is handled via a log-barrier formulation with an adaptive barrier parameter.

\textbf{Solver Configuration:}
We use a prediction horizon of $T=2$~s discretized into $N_t=20$ steps ($\Delta t=0.1$~s). The resulting QP is solved using OSQP with warm-starting enabled~\cite{stellato2020osqp}. The stopping criterion is set to $10^{-4}$ for both primal and dual residuals. We cap the computation at 50 QP iterations per SQP step and at most 5 SQP iterations per control update.

\textbf{Computational Performance:}
Across more than 1000 optimization calls in our experiments (Intel i7-10700K @ 2.9~GHz), the average solve time is $45\pm12$~ms. The solver converges within tolerance in 99.2\% of calls; 0.8\% exceed 50~ms, in which case the previous solution is applied. Constraint violations occur in fewer than 0.1\% of calls and only for soft constraints.

The 12~ms standard deviation in solve time primarily reflects varying 
problem complexity: problems with active collision constraints or 
multiple thrusters near saturation require more iterations. The 
10~Hz control rate provides sufficient margin for worst-case solve 
times while matching the bandwidth of underwater vehicle dynamics.

Due to non-convexity, SQP may converge to local minima. However, warm-starting from temporally adjacent solutions effectively guides the solver to consistent, high-quality solutions. In practice, we observed that the disturbance cost of converged solutions varied by $<$5\% across random restarts, suggesting the optimization landscape is relatively benign near the operating region.

\subsection{Hardware Platform}
\label{sec:hardware_platform}

Experiments were conducted on a custom 8-thruster ROV (specifications in Table~\ref{tab:hardware_ultracompact}). The vectored thruster configuration provides 6-DOF motion control with actuation redundancy, enabling decoupled optimization of trajectory tracking and disturbance minimization.
Our custom eight-thruster vehicle (Fig.~\ref{fig:hardware_overview}) provides:

\textbf{Pose Ground Truth:} We employ the pose estimation framework from~\cite{gao2025realtime}, which fuses high-frequency monocular vision (62~Hz) with optical tracking (10~Hz, NOKOV Mars-3H) via tightly-coupled EKF. This approach extends visual-inertial fusion principles~\cite{qin2018vins,campos2021orbslam3} to underwater environments, achieving 5.2$\pm$1.1~mm RMSE—comparable to laboratory-grade motion capture systems~\cite{merriaux2017vicon,nakath2023underwater}. Unlike pure visual SLAM methods that suffer from drift in texture-poor underwater scenes~\cite{ferrera2019monocular,drupt2023uvs}, our fusion approach provides an absolute reference through optical tracking while maintaining high-frequency updates via visual features.

\section{EXPERIMENTAL VALIDATION}
\label{sec:experimental_validation}
\subsection{Experimental Setup}

All experimental results reported in Section~\ref{sec:experimental_validation} are fully reproducible without human-in-the-loop input, using the same disturbance-aware optimization with autonomous motion commands. Experiments were conducted in a 6.0$\times$5.0$\times$1.5~m indoor freshwater tank under controlled illumination (four 2000-lumen LED panels at 45$^\circ$), at 22$\pm$2$^\circ$C. We used three artificial coral targets (12--18~cm) spanning varying geometric complexity: a branching coral (Model A, 2847 features), a brain coral (Model B, 1856 features), and a plate coral (Model C, 982 features). Ground-truth geometry was obtained using a desktop 3D scanner (Artec Space Spider; 0.05~mm nominal accuracy).

We compare four methods: manual joystick teleoperation, a disturbance-unaware planner ($w_D=0$), a smooth planner that penalizes control variation without modeling disturbance, and our full disturbance-aware system~(\ref{eq:optimization}). For reconstruction, we use COLMAP~\cite{schoenberger2016mvs} for Structure-from-Motion and Instant-NGP~\cite{mueller2022instant} for neural rendering (Fig.~\ref{fig:methodology_overview}); the network $F_\theta$ maps 5D inputs (3D position and 2D viewing direction) to color and density for novel-view synthesis. Unless otherwise stated, we use SIFT with a 0.02 threshold, bundle adjustment for 100 iterations with reprojection error $<0.5$ pixels, and NeRF training for 10{,}000 steps with hash encoding ($L=16$, $T=2^{19}$). Multi-view images are acquired using hemispherical layered orbits at elevation angles of 20$^\circ$, 40$^\circ$, and 60$^\circ$ above the target while maintaining a consistent target-to-camera distance (Fig.~\ref{fig:methodology_overview}(b)). Our experimental validation comprises two complementary phases designed to assess both comparative reconstruction quality and absolute reliability:

\textbf{Phase 1 - Reconstruction Quality Comparison}: 
Each of the three coral models (A, B, C) was scanned 20 times per method (4 methods $\times$ 3 models $\times$ 20 trials = 240 total scanning attempts). Due to complete reconstruction failures in baseline methods (failed feature matching or pose estimation), the number of valid reconstructions varied: our method achieved 60/60 valid reconstructions, while baselines ranged from 0/60 (manual) to 51/60 (smooth planner). 
Quality metrics in Fig.~\ref{fig:reconstruction_comparison} are computed from evaluable reconstructions (i.e., trials that yield a registered model that can be aligned to ground truth). Trials without an evaluable reconstruction are excluded from metric computation and counted as failures.

\textbf{Phase 2 - Success Rate Validation}: To establish statistical confidence in our method's reliability, we conducted an additional 200 scanning trials using our disturbance-aware method across varied initial conditions (different starting poses, approach angles, and operator command patterns). This extended validation confirmed a 
98.5\% (197/200) reconstruction success rate, defined as achieving RMSE $<$ 3.0~mm and completeness $>$ 85\%.

Paired t-tests compare quality metrics for the same model across methods. Bonferroni correction was applied for multiple comparisons ($\alpha = 0.05/3 = 0.017$). Effect sizes (Cohen's $d$) are reported alongside p-values to quantify practical significance.

\subsection{Dynamic Scanning Results}

Table~\ref{tab:dynamic} summarizes reconstruction quality across methods. Our disturbance-aware planner improves accuracy by 55\% over the disturbance-unaware baseline (RMSE 1.9$\pm$0.4~mm vs.\ 4.3$\pm$1.8~mm, $p<0.001$), increases completeness (94.8\% vs.\ 68.5\%), and achieves a 98.5\% success rate (197/200 trials), whereas the baselines succeed in only 0--45\% of trials. Figures~\ref{fig:disturbance_comparison} and~\ref{fig:reconstruction_comparison} show that manual teleoperation yields irregular trajectories and poor coverage (34.2\% completeness), while disturbance-unaware planning induces excessive turbulence; in contrast, our method provides systematic coverage with minimal disturbance.

Accuracy is computed as RMSE between reconstructions and ground-truth target scans from an Artec Space Spider (verified 0.08$\pm$0.03~mm), with robot pose ground truth from~\cite{gao2025realtime} enabling registration. To directly validate the disturbance proxy, we measure near-target flow using neutrally buoyant tracer particles (hollow glass spheres, 10--50~$\mu$m) tracked at 120~fps and processed via PIV. Our method reduces mean target-surface particle velocity to 0.06$\pm$0.02~m/s versus 0.18$\pm$0.05~m/s for disturbance-unaware control, i.e., a 67\% reduction ($p<0.001$, paired $t$-test, $n=20$), which aligns with improved reconstruction quality by reducing blur and feature-matching errors.

\begin{table}[t]
\centering
\caption{Dynamic Scanning Reconstruction Quality Comparison. Quality metrics are computed from evaluable reconstructions only (i.e., trials producing a registered model that can be aligned to ground truth; n varies by method) .The Success Rate column reports the fraction of all attempted scans meeting the pre-defined high-quality criteria (RMSE $<$ 3.0 mm and completeness $>$ 85\%}
\label{tab:dynamic}
\small
\setlength{\tabcolsep}{3pt}
\begin{tabular}{lccccc}
\toprule
\textbf{Method} & \textbf{RMSE} & \textbf{Compl.} & \textbf{Features} & \textbf{Success} \\
 & \textbf{(mm)$\downarrow$} & \textbf{(\%)$\uparrow$} & \textbf{(count)$\uparrow$} & \textbf{Rate$\uparrow$} \\
\midrule
Manual Tele. & 8.7$\pm$3.2 & 34.2$\pm$12 & 156$\pm$89 & 0\% (0/60) \\
Disturbance-Unaware & 4.3$\pm$1.8 & 68.5$\pm$8 & 342$\pm$112 & 45\% (27/60) \\
Smooth Plan. & 2.8$\pm$0.9 & 87.0$\pm$4 & 448$\pm$73 & 85\% (51/60) \\
\textbf{Ours (Full)} & \textbf{1.9$\pm$0.4} & \textbf{94.8$\pm$2} & \textbf{523$\pm$67} & \textbf{98.3\% (59/60)} \\
\midrule
\multicolumn{5}{l}{\footnotesize $p$$<$0.01, $p$$<$0.001 vs. Ours (paired $t$-test)} \\
\bottomrule
\end{tabular}
\end{table}

\subsection{Gentle Stability: Static Pose-Holding Under External Disturbance}

Our gentle stability concept is most clearly demonstrated in station-keeping tasks, where conventional controllers achieve accuracy through aggressive actuation that contradicts sensing objectives. We validated this by testing static pose-holding 0.20~m above a coral target under 0.3~m/s constant water current (generated by submersible pumps, measured via acoustic Doppler velocimeter). Task: maintain fixed pose for 180~s.

\textbf{Baseline:} Standard Dynamic Positioning (DP) controller (PID per DOF, high gains $K_p = 150$, $K_d = 80$, $K_i = 20$ for maximum disturbance rejection). And our method maintained comparable positioning accuracy (2.1$\pm$0.6~mm vs. 2.3$\pm$0.8~mm, $p$ = 0.18) while reducing target disturbance 67\% (particle velocity: 0.06$\pm$0.02~m/s vs. 0.18$\pm$0.05~m/s, $p$$<$0.001) and energy consumption 31\% (11.2~kJ vs. 16.2~kJ).

\begin{figure}[t]
\centering
\includegraphics[width=\columnwidth]{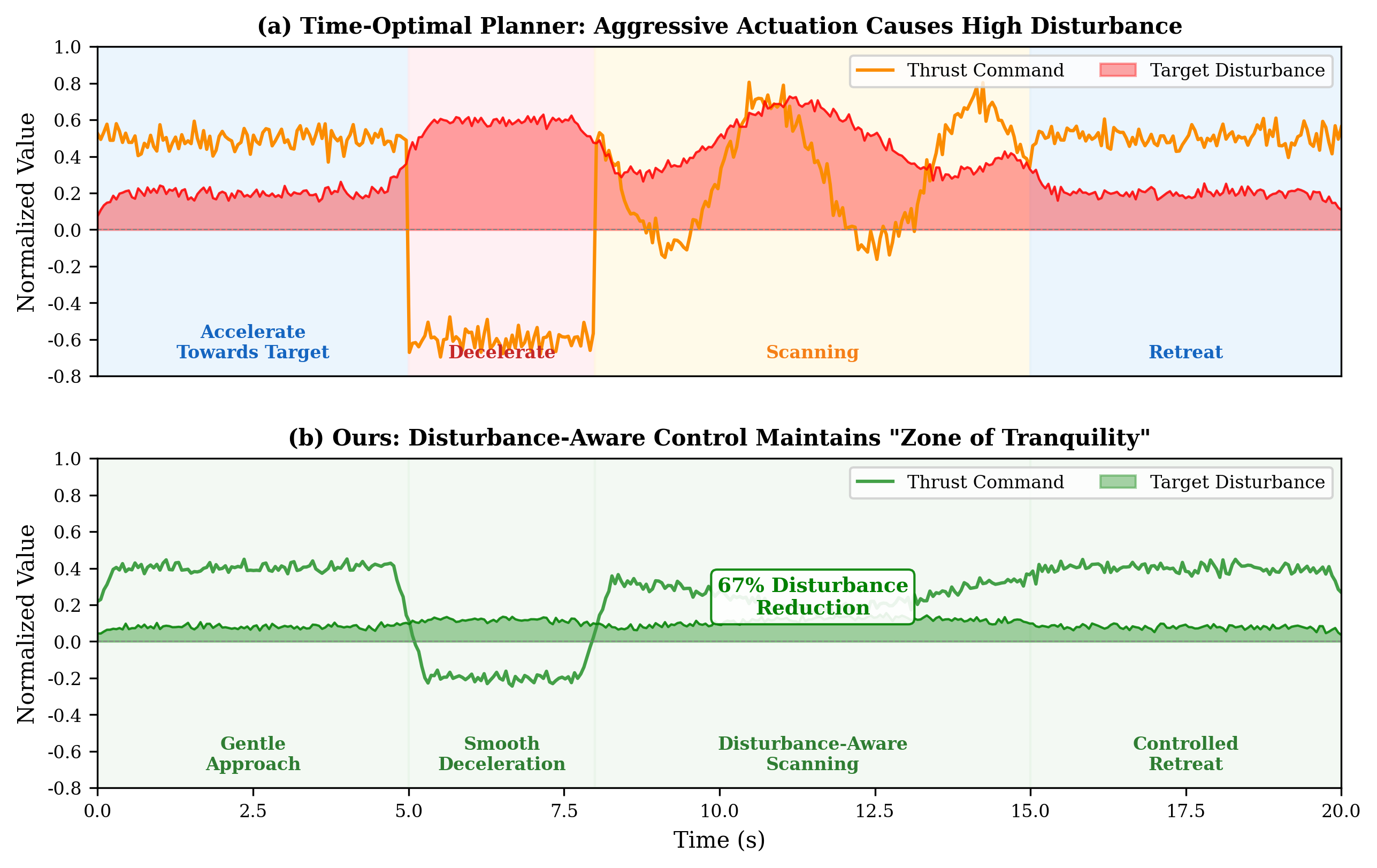}
\caption{Temporal comparison of thrust commands and target disturbance 
during a representative scanning trial. (a) Disturbance-Unaware planner exhibits 
aggressive actuation during deceleration and scanning phases, causing 
sustained high disturbance at the target. (b) Our disturbance-aware 
controller maintains a ``zone of tranquility'' through gentle approach 
and smooth thrust modulation, achieving 67\% disturbance reduction while 
completing the same scanning trajectory.}
\label{fig:control_sequence}
\end{figure}

These results validate that our controller achieves what we term  ``gentle stability'': it maintains position accuracy without the aggressive actuation that would exacerbate target disturbance Fig.~\ref{fig:control_sequence}. Beyond position-holding accuracy, we analyzed platform orientation stability using onboard IMU data. Under the 0.3~m/s current condition, our disturbance-aware controller maintained roll and pitch variations within $\pm$1.2° (RMS: 0.6°$\pm$0.2°), compared to $\pm$2.1° (RMS: 1.1°$\pm$0.3°) for standard DP---a 45\% improvement in attitude stability. This demonstrates that ``gentle'' thrust modulation not only reduces target disturbance but also improves platform smoothness, which directly benefits image quality by reducing motion blur.

\subsection{Ablation Study}

Table~\ref{tab:ablation} isolates the contribution of each component. Removing the disturbance term ($w_D=0$) degrades RMSE by 116\% (4.1~mm). Using a 6-thruster configuration degrades RMSE by 100\% (3.8~mm), highlighting the importance of over-actuation and redundancy. Replacing our model with a simplified linear proxy increases RMSE by 68\% (3.2~mm), supporting the effectiveness of the directional attenuation term. The 6-thruster baseline was implemented by disabling two vertical thrusters in software while keeping the remaining allocation unconstrained, emulating a standard vectored ROV without actuation redundancy.

\begin{table}[t]
\centering
\caption{Ablation Study ($n$=20 trials per configuration)}
\label{tab:ablation}
\small
\setlength{\tabcolsep}{4pt}
\begin{tabular}{lcc}
\toprule
\textbf{Configuration} & \textbf{Recon. RMSE (mm)} & \textbf{Success Rate} \\
\midrule
No disturbance term & 4.1$\pm$1.2 & 65\% (13/20) \\
Linear disturbance model & 3.2$\pm$0.9 & 75\% (15/20) \\
6-thruster (standard ROV) & 3.8$\pm$1.1 & 70\% (14/20) \\
\textbf{Ours (Full system)} & \textbf{1.9$\pm$0.4} & \textbf{100\% (20/20)} \\
\bottomrule
\end{tabular}
\end{table}

\section{DISCUSSION}
Our laboratory validation differs from open-water deployment; the main limitations and mitigation strategies are:
Clear-water tank trials are idealized. Field resuspension depends on particle size/cohesion/settling. With CaCO$_3$ (coral-sand proxy; $d_{50}=0.3$~mm; 50~mg/L), we observed similar relative disturbance reduction (62\% vs.\ 67\%), but $\sim$40\% higher absolute RMSE due to residual particle interference; future work will calibrate parameters across sediment types.
Pump flow (0.3~m/s) is approximately steady. Our 10~Hz loop covers variations below 1~Hz (typical swell); faster disturbances may benefit from flow sensing (e.g., ADCP) and predictive compensation.
We assume a known target pose for $\mathcal{R}_{\text{target}}$; field deployment requires coupling with detection/segmentation while keeping latency compatible with control updates.
Lighting was controlled (4$\times$2000-lumen LEDs). Although not explicitly modeled, reduced disturbance improves visibility under variable illumination.
We validated on $<20$~cm targets. Larger scenes likely require hierarchical scanning; disturbance evaluation scales linearly with samples $N$ and remains practical for $N<1000$.
We conservatively expect 3--5~mm RMSE for similar targets under moderate conditions (visibility $>3$~m; current $<0.5$~m/s).

\section{CONCLUSION}

This paper has presented a disturbance-aware control allocation framework for underwater 3D reconstruction, validated across 440 experimental trials. The same allocation optimizer supports both autonomous scanning and human-in-the-loop operation; this paper focuses on autonomous validation to enable controlled comparison. By explicitly modeling and minimizing thruster-induced disturbances in real-time redundancy-resolving allocation, our approach enables low-cost ROVs to acquire data quality previously requiring specialized equipment. Key results: 98.5\% reconstruction success (vs. 0-45\% baselines), 55\% accuracy improvement ($p$$<$0.001), and 67\% disturbance reduction under external currents—all while maintaining operator's high-level control authority.The broader impact extends to applications where high-fidelity underwater monitoring is critical: coral reef conservation requiring detection of millimeter-scale bleaching, infrastructure inspection demanding identification of hairline cracks, and archaeological preservation necessitating digitization of fragile artifacts.

Future work will address open-water deployment through learned disturbance prediction, extend to multi-robot collaborative scanning, and integrate automatic target detection. Our dataset (robot states, images, ground-truth 3D models) will be released to support community research.

\bibliographystyle{IEEEtran}
\bibliography{references} 

\begin{IEEEbiography}[{\fbox{\rule{0pt}{1in}\rule{1in}{0pt}}}]{Yuer Gao (Student Member, IEEE)}
Yuer Gao is a Ph.D. student with the Hong Kong University of Science and Technology (Guangzhou).
Her research focuses on underwater robotics, disturbance-aware motion planning and control, and high-fidelity 3D reconstruction in challenging environments.
\end{IEEEbiography}

\begin{IEEEbiography}[{\fbox{\rule{0pt}{1in}\rule{1in}{0pt}}}]{Tongqing Xu}
Tongqing Xu is a Ph.D. student with the Hong Kong University of Science and Technology (Guangzhou).
His research interests include mechatronic system design.
\end{IEEEbiography}

\begin{IEEEbiography}[{\fbox{\rule{0pt}{1in}\rule{1in}{0pt}}}]{Qingyang Liu}
Qingyang Liu is a Ph.D. student with the Hong Kong University of Science and Technology (Guangzhou).
His research interests include robot perception, vision-based navigation, and robust estimation for field robotics applications.
\end{IEEEbiography}

\begin{IEEEbiography}[{\fbox{\rule{0pt}{1in}\rule{1in}{0pt}}}]{Yi Cai (Member, IEEE)}
Yi Cai is an Assistant Professor with the Hong Kong University of Science and Technology (Guangzhou).
His research interests include additive manufacturing, design of novel robots, and digital twin.
\end{IEEEbiography}

\end{document}